# Ensemble Convolutional Neural Networks for Mode Inference in Smartphone Travel Survey

Ali Yazdizadeh, Department of Geography, Planning and Environment, Concordia University, Canada.
Zachary Patterson, Department of Geography, Planning and Environment, Concordia University, Canada.
Bilal Farooq, Department of Civil Engineering, Ryerson University, Canada.

*Abstract*—We develop ensemble Convolutional Neural Networks (CNNs) to classify the transportation mode of trip data collected as part of a large-scale smartphone travel survey in Montreal, Canada. Our proposed ensemble library is composed of a series of CNN models with different hyper-parameter values and CNN architectures. In our final model, we combine the output of CNN models using "average voting", "majority voting" and "optimal weights" methods. Furthermore, we exploit the ensemble library by deploying a Random Forest model as a meta-learner. The ensemble method with random forest as meta-learner shows an accuracy of 91.8% which surpasses the other three ensemble combination methods, as well as other comparable models reported in the literature. The "majority voting" and "optimal weights" combination methods result in prediction accuracy rates around 89%, while "average voting" is able to achieve an accuracy of only 85%.

*Index Terms*—Deep learning, convolutional neural networks, meta learning, smartphone travel surveys, GPS trajectories, mode inference, machine learning

## I. Introduction

In the last two decades, advances in data collection techniques, particularly in the use of new technologies such as Global Positioning Systems (GPS) and smartphones have been changing the options available for collecting transportation demand data. Using the data collected from these emerging technologies typically requires methods of data processing and inference. Detecting transport mode has received the lion's share of attention in the literature as researchers have tried to infer mode with methods ranging from rule-based approaches to artificial intelligence algorithms.

In recent years, deep neural networks have achieved considerable success in various applications, particularly in image recognition tasks. The performance of traditional machine learning methods are highly dependent on the choice of data representation (or features). Also a major part of efforts in applying them is spent on pre-processing data and extracting "hand-crafted" features [1]. This process of feature engineering is not only labor-intensive, but also underlines the weakness of many machine learning algorithms; their lack of ability for them to derive important factors from data [1], [2]. Moreover traditional approaches to feature engineering are highly dependent upon prior or commonsense knowledge. Deep Learning approaches are increasingly used to automatically engineer features for use in machine learning algorithms since they can uncover multiple levels of representation, with higher-level features representing more abstract aspects of underlying datasets [1].

With respect to inferring transportation mode with machine learning algorithms, hand-crafted aggregate trip features such as trip length, mean speed, mean acceleration, etc. are typically provided to classifier algorithms. As such, these methods typically fail to take account of the potentially rich data available on individual GPS points along a trip. In addition, some engineered features, such as maximum speed or average acceleration can be strongly collinear making it difficult for classifiers to use them to distinguish between modes. Deep learning approaches allow for the automated discovery of abstraction that can not only reduce the dependence of mode inference algorithms on feature engineering, or on prior knowledge of the modeller, but can also discover factors related to each GPS point that are usually overlooked by traditional machine learning algorithms. In this study, we use GPS-point information as input for Convolutional Neural Networks (CNN) to infer the mode of transport. In addition, we take advantage of ensemble learning strategies to improve the prediction accuracy of CNN models.

This research uses data collected by the smartphone travel survey app, MTL Trajet, which is an instance of the smartphone travel survey platform, DataMobile/Itinerum [3]. MTL Trajet was released as part of a large-scale pilot study on the 17[th] of October 2016 in a study that lasted 30 days. Over 8,000 people participated in the study [4].

The rest of the paper is organized as follows: a background section describes previous work on transport mode inference. The methodology section sets out the framework of the CNN as well as of the data pre-processing and ensemble method configuration. The next section after that presents the results of the CNN and ensemble models on the MTL Trajet dataset. The last section concludes the paper.

## II. Background

This section reviews previous research related to mode detection from smartphone data. We also describe the mathematical operations used in the convolutional neural networks (CNN).

### A. Mode Detection and Machine Learning

Mode detection has been done using various approaches in the literature, which include: rule-based, machine learning, and discrete choice approaches. Elhoushi et al.[5] have done a comprehensive and comparative literature review on travel mode detection using different sensors and different classification methods.

Tree-based ensemble classification algorithms have been used by Xiao et al. [6] to classify mode of transport. Their best ensemble method achieved a prediction accuracy of 90.77%. Wang et al. [7] develop a Random Forest classifier combined with a rule-based method to detect six modes of transportation using seven GPS-related features. Their method is able to detect more than 98% of subway trips with an overall accuracy of the other five modes classification as high as 93.11%. Assemi et al. [8] deployed a nested logit model with eight attributes to infer mode of transport from smartphone travel surveys, implemented in New Zealand and Australia. They have reported an accuracy of 97% for New Zealand which includes data cleaning and 79.3% for Australia without any pre-processing.

Endo et al. [9] proposed a deep neural network approach to automatically extract high-level features. Their innovative approach converted raw GPS trajectories into a 2-D image structure and fed it as the input to a deep neural network. As an alternative to the RGB (Red, Green, Blue) values of an image pixel, stay time, i.e. the duration that a user stays in the location of the pixel, was used as the pixel value. They integrated hand-crafted features with the image-based feature. Eventually, they deployed traditional machine learning models, such as logistic regression and decision tree, to predict mode of transport. Although they devised an innovative idea to convert GPS trajectories into 2-D images, the pixel values only contained stay time without taking into account the spatiotemporal or motion characteristics, such as speed or acceleration, of the GPS trajectories. Their best models was able to detect the mode of transport with prediction accuracy of 67.9% on the GeoLife dataset and 83.2% on the Kanto Trajectories dataset.

More recently, Dabiri and Heaslip [10] used CNN models to train a mode detection classifier. They developed different architectures of CNN models on GPS trajectories, and finally combined their output via an ensemble method. Their ensemble library comprised seven CNN models. They took the average of the softmax class probabilities, predicted by each CNN model to generate the transportation label posteriors. Although the study carried out by Dabiri and Heaslip [10] used the CNN models, their study is different from our study in the following ways.

The segmentation method; the CNN architectures (explained in Section III) and the survey size used in two studies are different. The latter is particularly important. The current study is based on a large-scale, real-world travel survey with about 8,000 participants, while the Dabiri and Heaslip study used trajectory data from 69 users. Furthermore, we have used a different ensemble method, i.e. a random forest model as a meta-learner explained in Section III, which demonstrates better prediction performance over the ensemble method developed by Dabiri and Heaslip [10].

Apart from the features used in the mode inference literature, features used in other fields of study may contribute to transportation mode detection [10]. "Jerk" has been used in traffic safety analysis to identify critical traffic events. Jerk is defined as the derivative of acceleration, or the rate of change of acceleration over time[10]. GPS bearing rate has been used in driver behaviour profiling using smartphone data [11]. Bearing rate is defined as the change of bearing between three consecutive GPS points [10] (Jerk and bearing rate are further explained in Section III-A).

With respect to Convolutional Neural Networks, numerous architectures have been implemented in the broad Deep Learning literature, although their fundamental elements are very much alike. With respect to data dimensions, CNNs have been implemented on 1D, 2D and 3D data [2]. Since CNNs can learn features as well as estimate classifier coefficients, they are able to accomplish better prediction accuracy on large-scale datasets [2]. The success of CNNs in other fields of study encourages us to implement it as a classifier to infer mode of transport. The data used in this study represents a sequence of GPS points for each trip and can be considered as one-dimensional data. However, since CNNs typically use same-size input, we split the trip trajectories into same-length segments. The data preparation steps and modelling approaches have been explained in the next section.

### B. Convolutional Neural Networks

CNN models are a class of neural networks suitable for processing grid-like topology data [12], which vary from 1D time-series data to 2D images. CNN models rely on affine transformation [12], which involves a vector of inputs being multiplied by a matrix (also called kernels, or filters) to produce an output. The multiplication by a matrix is referred to as convolution operation. Typically, a bias vector is added to the result of the matrix multiplication. Next, a non-linear function, called an activation function, is applied to the output of aforementioned operations. After the non-linear activation function, a pooling operation is typically applied. We briefly explain each of these stages below.

Generally, these mathematical operations, i.e. matrix multiplication, function activation and pooling, form one "hidden layer" of a CNN model. The output of each hidden layer of a CNN model can be fed as input into the next layer. The last layer of a CNN model produces class probabilities by applying an activation function, such as the sigmoid or softmax functions. Figure1 shows the general architecture of a CNN model.

Regardless of the dimensionality, the input data for a CNN model can be stored as fixed-sized multi-dimensional arrays (or tensors) [13]. Hence, GPS trajectories in our dataset need to be converted to fixed-sized arrays with different channels. This procedure is explained in SectionIII-A.

### C. Ensemble methods

Ensemble methods combine multiple classifiers and have been found to provide the possibility of higher accuracy results than a single classifier. Well-known ensemble techniques include boosting, bagging and stacking. Stacking combines the outputs of a set of base learners and lets another algorithm, referred to as the meta-learner, make the final predictions [14]. A super learner is another method that calculates the final predictions by finding the optimal weights of the base learners by minimizing a loss function based on the cross-validated output of the learners [14].

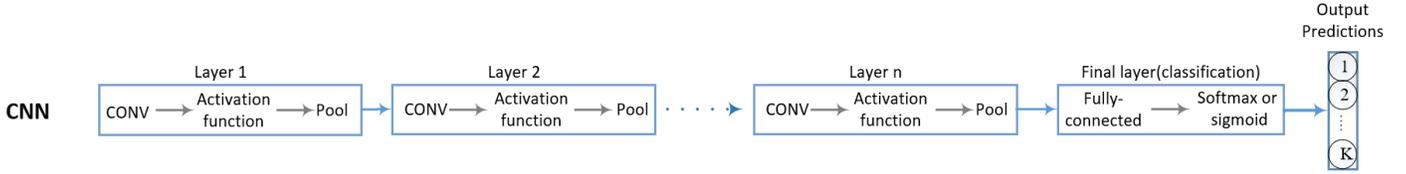

Fig. 1. General architecture of convolutional neural network. (CONV: Convolution)

The most common ensemble method used for neural networks is average voting that generates posterior labels by calculating the average of the softmax class probabilities or predicted labels for all the base learners [14]. Majority voting is another ensemble approach that counts the predicted labels from all the base learners and reports the label with the maximum number of votes as the final prediction. Another approach is to calculate the optimal weights of individual base learners. The optimal weight of each base learner can be obtained by minimizing a loss function, i.e. the mean square error (MSE), given output of the base-learners. By minimizing the loss function, better performing classifiers are assigned larger weights. The final predictions of the ensemble are acquired by voting using the optimal weights of classifiers. Using a meta-learner is another ensemble approach that trains a learner, e.g. a Random Forest model, on the predicted class probabilities (or labels) of all base-learners to make the final prediction.

III. METHODOLOGY

In this section, we describe the data used, data pre-processing, the CNN architectures used, hyper-parameter value determination, and ensemble model configuration.

*A. Data*

Three types of data from the MTL Trajet dataset were used in this analysis; trip mode of frequent trips, coordinates data and mode prompt data. Upon installation of MTL Trajet, respondents were asked a series of questions. Two of these questions related to travel mode for trips from home to work or study location. For trips between these locations, respondents were asked what mode of travel they typically used, and whether or not other modes were ever used. Coordinates data contain respondent latitude and longitude obtained primarily through GPS. There are over 33 million location (primarily GPS) points in the MTL Trajet 2016 database. Mode prompt data include information on trip mode, i.e. walk, bike, car, public transit, provided by respondents. When analyzing the data, it was found that mode data for trips between home, work and study for respondents that used one (and only one) mode was less noisy than mode data recorded from prompts. Hence, only validated trips from users who declared they used only one mode option to travel between home and work or home and school were used.

We used the rule-based trip-breaking algorithm developed in Patterson & Fitzsimmons [3] to detect start and end point of trips from raw GPS trajectories. The trip-breaking algorithm considered dwell time between GPS points as the most prominent criterion for detecting trips. The trip-breaking algorithm detects trip segments based on 3-min gaps in data. Segments are then stitched back together while controlling for velocity and parameters relating to the public transit network (i.e. transit junctions and metro station location). For example, when two consecutive points are detected within 300m of a metro station (data collection is sparse when underground), and the gap is less than the maximum travel time by metro, the segments are joined as part of the same trip. Similarly, when two consecutive points fall within the same intersection with bus correspondences, additional time (10 min instead of 3 min gap) was allowed.

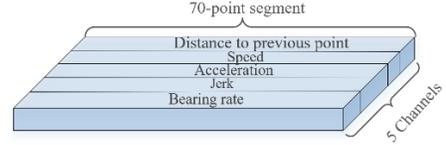

Fig. 2. A 70-point Segment with 5 Channels.

*B. Data Preparation*

We considered five channels for the input data for the CNN models: distance to previous point, speed, acceleration, jerk, and bearing rate, as recommended in the literature [10]. Speed is calculated using the distance between each two consecutive GPS points divided by their time interval. Acceleration is also defined as the derivative of speed or the rate of change of speed over time.

These characteristics constitute the channels of each segment as shown in Figure 2. These segments are fed as input to the CNN models. Jerk is defined as the rate of change in acceleration. Bearing is a term used typically in navigation and defined as the angle between the direction of vehicle to the destination and the magnetic north. Bearing is different from heading, as heading is the direction in which a vehicle is moving. However, like bearing, the heading is defined in degrees from magnetic north. Figure 3 demonstrates the bearing rate of a moving vehicle.

Bearing rate values vary across different transportation modes. For example, buses and cars usually do not experience sharp changes in steering angle while travelling between traffic lanes. As a result, their heading does not change regularly. However, pedestrians and bike riders more commonly experience changes in their heading. The formula to calculate the bearing of two consecutive GPS points ($p_1$, $p_2$) is defined by following equations [10]:

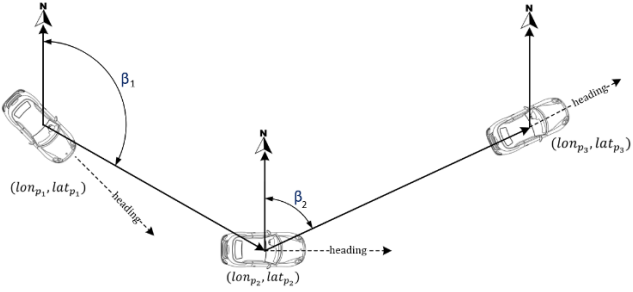

Fig. 3. Bearing ($\beta_1$ and $\beta_2$) and Heading of a Moving Vehicle.

$$\beta_1 = \arctan(X, Y) \quad (1)$$

where:

$$X = \cos(lat_{p_1}) * \sin(lat_{p_2}) - \sin(lat_{p_1}) * \cos(lat_{p_2}) * \cos(lon_{p_2} - lon_{p_1})$$
$$Y = \sin(lon_{p_2} - lon_{p_1}) * \cos(lat_{p_2})$$

The *lon* and *lat* represents the longitude and latitude of GPS points in radians. The bearing rate is defined as the absolute change in the bearing of two consecutive GPS points and can be obtained by the following formula [10]:

$$Bearing\ Rate = |\beta_2 - \beta_1| \quad (2)$$

As shown in Figure 3, at least three consecutive GPS points are required to calculate the bearing rate, since the bearing of a GPS point is calculated according to its following point. For example, as shown in Figure 3 the bearing of vehicle at the first position, i.e. $P_1$, is calculated according to $P_2$, and the bearing of the vehicle at $P_2$ is calculated according to $P_3$, as in above formulas.

The input data of the CNN models needs to be of fixed size. As the number of GPS points along trips vary, we tried to split trips into m-point segments, where m was the average (70 points) or median (120 points) number of points for trips in the dataset. After examining both 70 and 120 point segments we found the best performance of CNN with seventy-point (average length) segments. For those segments with the number of points less than 70, the remainder of the segment was padded with zeroes. The resulting dataset consisted of 3,845, 8,515, 7,415 and 15,275 walk, bike, transit and car segments, respectively.

We also implemented other data pre-processing steps to remove errors from GPS trajectories. We removed trajectories with less than 10 GPS points. Furthermore, we only considered segments from those respondents who had validated at least three trips during the survey. Also, restrictions on speed and acceleration between two consecutive GPS points was applied as suggested by Dabiri and Heaslip[10].

### C. CNN Architecture

The overall architecture of the CNNs used in the current study is illustrated in Figure 4. A typical CNN architecture consists of three layers: convolutional, pooling and fully-connected layers. Our model begins with segments of size $70 \times 5$ that contains the 70 GPS points and 5 channels. The convolutional layers of the CNN are formed by a convolution operation followed by a max-pooling operation. The Leaky Rectified Linear Unit (Leaky ReLU) activation function is applied to the output of every convolutional layer. The last layer is fully-connected. The output of the last fully-connected layer is fed to a 4-way softmax that produces a distribution over the 4 modes of transportation.

Seven different CNN architectures, shown in TableI, were evaluated. Model A is a basic model with only one convolutional and one fully-connected layer. Model B consists of four convolutional and one fully-connected layer with convolutional and pooling kernels of size $8 \times 1$. Model C uses a deeper architecture consisting of 20 convolutional layers and one fully connected layer. It has the same number of kernels and kernel sizes for convolution as Model B, but with a max-pooling layer of size $4 \times 1$. We make a deeper architecture in Model C by repeating the layers five times. We implement Model C (and also Model E, as explained below) to evaluate the performance of wide versus deep architectures for CNN models on our dataset.

Model D is a wide neural network, i.e. it uses a large number of convolutional kernels relative to the number of layers. The architecture consists of 5 convolutional layers followed by one fully connected layer. The number of kernels in each layer is similar to the ones suggested by Krizhevsky [15]. The first convolutional layer consists of 96 kernels of size $8 \times 5$, which are applied with a stride of 1 on the $70 \times 5$ input segment. The second convolutional layer takes the output of the first convolutional layer as input and has 256 kernels of size $8 \times 96$. There are 384 kernels of size $8 \times 256$ in the third and 384 kernels of size $8 \times 384$ in the fourth convolutional layer. Also, the fifth convolutional layer consists of 256 kernels of size $8 \times 384$. Each convolutional layer is followed by a Leaky ReLU activation function and a max-pooling layer of size $8 \times 1$. To keep the output size the same as the input size we use the *same* padding for both convolutional and max-pooling kernels [13]. Also, we use stride 1 for both convolutional and max-pooling layers to allow a high spatial resolution of GPS points.

Model E uses a deeper architecture, relative to Model D, consisting of 20 convolutional layers and one fully connected layer. The architecture of Model E is similar to the architecture of Model D in terms of number of kernels and kernel sizes. To reach a deeper architecture, each layer is repeated four times. We use a $2 \times 1$ max-pooling layer after each convolutional layer with unit stride and the *same* padding. Model F contains a larger number of kernels than the previous models. We implement Model F to assess how far the larger number of kernels can improve the prediction accuracy of a CNN model. The model consists of 6 convolutional layers with 128, 256, 512, 1024, 1024 and 512 kernels. The convolutional and pooling kernels are of size $8 \times 1$. Finally, we applied the fully-connected layer. For all the models, to avoid over-fitting we applied the Dropout method before the fully-connected layer, and set each hidden neuron to zero with a probability of 0.5.

We tested different values for the size of convolutional

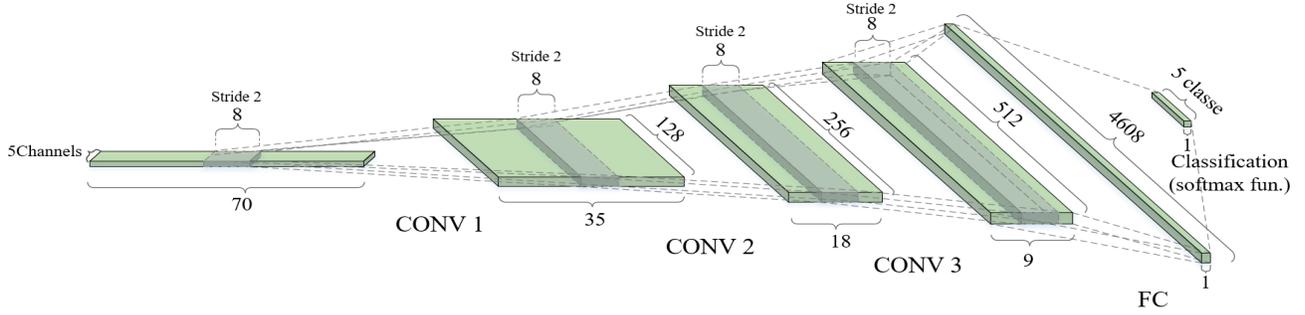

Fig. 4. Convolutional Neural Network Architecture with N convolution layers and one Fully-connected layer.

(kernel) and pooling size, from 2 to 16. We found lower prediction accuracies when the kernel or pooling sizes were set to 16. However, with respect to kernel size, all the models with kernel size of 8 showed higher prediction accuracy. This was not the same for pooling size, where as in deeper models, such as models B and E, the prediction accuracies were better when we set the pooling size to 4 in Model B and 2 in Model E.

*D. Ensemble Model*

In addition, we apply four methods to combine the predictions of CNN models: (a) average voting (b) majority voting (c) optimal weights of individual base learners and (d) a random-forest model as meta learner. For the third method, to find the optimal weight of each base-learner, we minimize a loss function, i.e. the mean square error (MSE), given output of the base-learners. By minimizing the loss function, better performing classifiers are assigned larger weights. The final predictions of the ensemble are acquired by voting using the optimal weights of classifiers.

With respect to the fourth method, the predictions of base learners are fed to a random forest classifier to make the final predictions. The random forest model was built using 800 trees. Also, we allowed up to 8 attributes to be randomly sampled at each split.

In this study a series of CNN models as well as an ensemble method have been used. We used ensemble models to improve the prediction accuracy of the CNN models. In addition, we tested four different ensemble approaches for achieving the highest prediction accuracy. We refer to the CNN models in the ensemble library as level-0 models and the methods to make the final predictions as level-1 models. Level-0 includes models A-F (Table I) as well as other base learners developed based on different CNN architecture types and hyper-parameter values, shown in Table II. Level-0 contains 75 base learners. Finally, each method in level-1 is applied on the output of Level-0 models to predict the final labels.

## IV. RESULTS

The CNN architectures were programmed in Tensorflow with GPU support. We used randomized stratified sampling to ensure that all modes were represented equally in the training and test sets.

*A. CNN models*

In our first experiment we assess different CNN architectures in terms of: number of layers and kernels, kernel and pooling size, stride size, activation function, optimization method and batch size. We tested both Gradient Descent and Adam optimization algorithms, and found that the Adam algorithm compares favorably to Gradient Descent. For Adam, we used a decay learning rate equal to 0.95 and set the starting learning rate to 0.01. We also tested the fixed learning rate equal to $1e-4$ and found there is not any significant difference between the results of models with decay or constant learning rates. Hence, we set the learning rate to $1e-4$ for all the models. We did our first experiments with number of epochs equal to 20 and 50. To avoid over-fitting we used dropout and early stopping methods [16]. With respect to batch size, we tested sizes of 16, 32, 64 and 128, and got better results with a batch size of 16. We also tested 'tanh', 'Relu' and 'Leaky Relu' activation functions and achieved better results with Leaky Relu function.

Table III shows the results of six CNN models as well as the ensemble models. Model A is a basic model and shows that a CNN with only one convolutional layer is able to predict the mode of transport with 72.5% accuracy. Models B and C have different numbers of layers, i.e. 6 and 21 layers respectively, and possess the same number of kernels as well as kernel and pooling size. The prediction accuracy of Models B and C is almost identical, although model C is marginally better than model B.

Models D-F comprise higher numbers of kernels in each layer than the previous models and display higher prediction accuracies. Models D-F show prediction accuracies of 81.3%, 80.6% and 80.1%, respectively. The trade-offs between the depth (number of layers) and width (number of kernels) of neural networks have been investigated by some researchers [17]. In fact, deeper networks can show better results than shallower networks [17]. However, as He et al. [17] have mentioned, at some point, the error in prediction accuracy of CNN models not only gets saturated, but gets worse as the models get deeper. We see such a pattern when comparing models E and F with model D that while deeper has lower prediction accuracy. Similar observations have been reported by Dabiri and Heaslip [10]; they also found that a shallower network performed better than a network with deeper configuration.

TABLE I
CNN MODEL ARCHITECTURES

| Model A | Model B | | Model C | Model D | | Model E | Model F |
|---|---|---|---|---|---|---|---|
| 2 layers | 5 layers | | 21 layers | 6 layers | | 21 layers | 7 Layers |
| Input segment | | | | | | | |
| conv 8-4 maxpool 4 | conv 8-4 maxpool 8 | $5 \times$ | $\begin{pmatrix} conv8-4 \\ maxpool4 \end{pmatrix}$ | conv 8-96 maxpool 8 | $4 \times$ | $\begin{pmatrix} conv8-96 \\ maxpool2 \end{pmatrix}$ | conv 8-128 maxpool8 |
| | conv 8-8 maxpool 8 | $5 \times$ | $\begin{pmatrix} conv8-8 \\ maxpool4 \end{pmatrix}$ | conv8-256 maxpool 8 | $4 \times$ | $\begin{pmatrix} conv8-256 \\ maxpool2 \end{pmatrix}$ | conv8-256 maxpool 8 |
| | conv 8-16 maxpool 8 | $5 \times$ | $\begin{pmatrix} conv8-16 \\ maxpool4 \end{pmatrix}$ | conv8-384 maxpool 8 | $4 \times$ | $\begin{pmatrix} conv8-384 \\ maxpool2 \end{pmatrix}$ | conv8-512 maxpool 8 |
| | conv 8-32 maxpool 8 | $5 \times$ | $\begin{pmatrix} conv8-32 \\ maxpool4 \end{pmatrix}$ | conv8-384 maxpool 8 | $4 \times$ | $\begin{pmatrix} conv8-384 \\ maxpool2 \end{pmatrix}$ | conv8-1024 maxpool 8 |
| | | | | conv8-256 maxpool 8 | $4 \times$ | $\begin{pmatrix} conv8-256 \\ maxpool2 \end{pmatrix}$ | conv8-1024 maxpool 8 |
| | | | | | | | conv8-512 maxpool 8 |
| Dropout | | | | | | | |
| Fully Connected | | | | | | | |

TABLE II
HYPER-PARAMETER VALUES USED IN ENSEMBLE

| Name | Hyper-parameter value |
|---|---|
| Convolutional layer | Num. of layers: 6, 7, 11, 21<br>Num. of kernels: 2,4,8,16,32,98,128,256,384,512,1024<br>Kernel size: 2, 4, 8<br>Stride size: 1, 2<br>Stride type:SAME<br>Activation function: Leaky ReLU |
| Max pooling layer | Pooling size: 2, 4, 8<br>Stride size: 1, 2<br>Stride type:SAME |
| Output layer | Activation: Softmax |
| Optimization method | Adam optimizer with learning rate = 1e-4 |
| Batch size | 16 |
| Number of epochs | 20, 50, 100 |

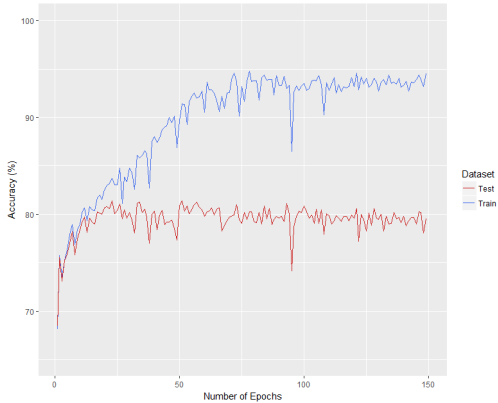

Fig. 5. Train and Test Accuracy for Different Numbers of Epochs (Model D).

### B. Ensemble model Results

Table III illustrates the prediction accuracy of ensemble models with different combining methods. The first method (average voting) leads to a higher accuracy than even the best CNN model, i.e. Model D. A better performance (89.4% prediction accuracy) is obtained using the majority voting method. Finding the optimal weights of base CNN models leads to the same prediction accuracy as the majority voting method. The highest prediction accuracy is obtained when using the Random Forest model as meta-learner (roughly 91.8%).

Table IV shows the confusion matrix of the ensemble method with RF as meta-learner, using a 5-fold cross validation. Walk, bike, transit and car segments are predicted with 91.1%, 91.8%, 84.8%, 95.7% precision, respectively. Recall and F-score are also reported with the highest and lowest values for car and transit. Car trips are easier to infer due to their higher speed and acceleration. However, bus trips share some similarities with car trips, especially in dense urban areas where cars and buses experience the same traffic conditions.

Figure 5 shows the train and test accuracy for Model D. The train accuracy increases with the number of epochs and the test accuracy remains constant after around 50 epochs. Apart from what was mentioned above, another reason that test accuracy in neural networks does not increase while the train accuracy goes higher without any over-fitting is the insufficient number of samples in the training data [10]. Hence, the inclusion of more transit trips to the dataset could possibly enhance the prediction accuracy of the CNN models.

### C. Comparison with classical machine learning models and previous studies

To assess the prediction accuracy of the CNN models and ensemble methods in this study, we developed a Decision Tree (DT) as a classical machine learning algorithm as well as a Random Forest (RF) model, which is an ensemble machine learning approach. We used several hand-crafted features, as explained in [18], including average speed (km/h), 85th percentile speed, maximum and minimum acceleration, travel time, etc. Like CNN models, there are hyper-parameters for DT and RF. The tree depth is controlled by the required minimum number of observations in each node. We tested

TABLE III
Test Accuracy of All Models

| Model type | Model A | Model B | Model C | Model D | Model E | Model F | Ensemble (Average Voting) | Ensemble (majority voting) | Ensemble (optimal weights) | Ensemble (RF as meta-learner) |
|---|---|---|---|---|---|---|---|---|---|---|
| Accuracy(%) | 72.5 | 75.6 | 75.7 | 81.3 | 80.6 | 80.1 | 85.0 | 89.1 | 89.4 | 91.8 |

TABLE IV
confusion matrix analysis (ensemble with RF as meta-learner)

| Mode of Transport | Walk | Bike | Transit | Car | Precision(%) | Recall (%) | F-score(%) |
|---|---|---|---|---|---|---|---|
| Walk | 708 | 32 | 29 | 0 | 91.1 | 92.0 | 91.6 |
| Bike | 32 | 1573 | 60 | 38 | 91.8 | 92.3 | 92.1 |
| Transit | 30 | 66 | 1296 | 91 | 84.8 | 87.4 | 86.1 |
| Car | 7 | 42 | 142 | 2864 | 95.7 | 93.8 | 94.7 |

different values, ranging from 1 to 40 for this parameter. The highest prediction accuracy on test datasets was acquired with the required minimum number of observations in each node equal to 20. With respect to the RF model we tried different numbers of trees in each forest, ranging from 100 to 2000. We found no improvement in prediction accuracy of RF models with number of trees higher than 1000. The resulting DT reached the prediction accuracy of 73.4% for detecting mode of transport. Also, the RF model, reached a predicting accuracy of 86.8%. Comparing these results with the results of CNN models and ensemble models shows that the ensemble methods, either a random forest (ensemble of decision trees) model or an ensemble of CNN models, outperform the single learners, such as the DT or single CNN models in our study. In addition, the CNN model in some cases, such as Models A-C in Table III, demonstrates a prediction accuracy close to the DT. However, with wider or deeper CNN architectures, such as Models D-F, we can reach higher accuracy prediction performance.

It has been said that comparing prediction accuracy rates from different studies can be misleading because of differences in sample sizes, number of classes, and the quality of data across studies. Bearing these considerations in mind, our findings compare favorably with previously obtained results in the literature. Dabiri and Heaslip [10] have reported a test accuracy of 84.8% for an ensemble convolutional neural network. Their best model, is able to predict walk, bike, bus, driving and train, with 81.6%, 90.3%, 80.7%, 86.6%, 92.3% precision, respectively. Also, Zheng et al. [19] have reported accuracies of 89%, 86%, 66% and 65% for walk, driving, bus and bike modes, respectively. Their decision tree-based inference model, results in an overall accuracy of 76.2%. Both the aforementioned studies [10], [19] used GPS data from mobile phones, similar to that collected in the MTL Trajet data, although they had much smaller sample sizes of 69 and 65 persons, respectively. Our best ensemble model shows an overall accuracy of 91.8% and is able to predict walk, bike, public transit, car with 91.1%, 91.8%, 85.8%, 95.7% precision based on data from over 8,000 users. The results demonstrate that overall accuracy of our best model is superior to the overall accuracy of both aforementioned comparable studies.

Also, Gonzalez et al. [20] produced precision rates of 92% for car trips and 81% for bus trips. Bantis and Haworth [21] developed several models to detect three modes of transport (walk, bus/car and train) using stationary points from GPS, accelerometer data as well as user characteristics. Their training data contained trips from 5 individuals. They proposed a hierarchical dynamic Bayesian network and compared the results against random forest, SVM, and Multilayer Perceptron classifiers. The accuracy of their proposed model was 90%, while it could predict Stationary status as well as Walk, Bus/Car and Rail trips with 94%, 65%, 88%, 91% precision, respectively. Although their categories are different from ours and the sample size of the two studies are not similar, our best ensemble model predicts walking trips with much higher precision, i.e. 91.1% vs. 65%. Moreover, our best ensemble model predicts car trips with 95.7 precision, higher than all aforementioned studies.

V. Conclusion

We developed a series of Convolutional Neural Networks augmented by different ensemble methods to infer travel mode from trajectories gathered by a large-scale smartphone travel survey. The raw trajectories of travellers were tailored in a manner that allowed feeding them as an input layer to a CNN. Each trip was segmented into fixed sized segments with five channels. The channels include "distance to previous point", "speed", "acceleration", "jerk", and "heading rate". We investigated different CNN architectures and combined their results via different ensemble methods to obtain the highest prediction accuracy. Our ensemble library was composed of a series of CNN models with different hyper-parameter values and CNN architectures. Afterwards, we combined the results of CNN models with "average voting," "majority voting" and predicting the optimal weight of each classifier. Furthermore, we exploited the ensemble library by deploying Random Forest as a meta-learner. We find that all the ensemble methods outperform the individual CNN models. Moreover, the ensemble method with random forest as meta-learner shows an accuracy of 91.8% which surpasses the other three methods. Finally, the "majority voting" and "optimal weight" combination methods result in similar prediction accuracy rates around 89%.

This study contributes to the mode detection literature and the ITS community by using a meta-learner classifier (random forest in our case) to aggregate the output of several base-learners (CNN models). While an approach using a meta learner ensemble, i.e. applying a learner over the output of a set of CNN base-learners, has been used in some applications in other fields [14], to the best of our knowledge the approach has not been used in previous mode detection studies. Furthermore, the data set used in this study has been collected as part of a large-scale travel survey, that shows the capability of such smartphone-based travel surveys as a complementary (or even a replacement) surveying tool to the current real-world household travel surveys.

In addition, considering that our study has been developed on trajectories gathered as part of a large-scale and real world travel survey, we believe that smartphone travel surveys have the potential to replace traditional travel surveys, such as face-to-face or computer assisted telephone interviewing(CATI), although there are still many hurdles to overcome. One of these hurdles is the quality of the labelled trajectories. The labelled trajectories produced by large-scale smartphone travel surveys may not show the same quality as labelled trajectories from small or researcher-collected smartphone travel surveys. Improving the trajectory collection and validation techniques for large-scale travel surveys should be a focus of future studies while deploying state-of-the-art classification techniques. Furthermore, as the need for labelled data is a requirement of supervised machine learning models used in the literature and the current study, using semi-supervised or unsupervised models may help to reduce or eliminate the need for labelled data.

### Acknowledgments

This research is based upon work supported by the Social Sciences and Humanities Research Council of Canada as well as "MTL Trajet" research contract with Ville de Montréal. Also, this research was enabled in part by support provided by WestGrid Computing Facilities and Compute Canada.

### Authors' Biography

**Ali Yazdizadeh** received Master in Transportation Planning from Sharif University of Technology, Iran, and Master in Geography and Planning from Concordia University, Canada in 2012 and 2016, respectively. He is currently working toward the Ph.D. degree in Geography and Planning at Transportation Research for Integrated Planning (TRIP) Lab, Concordia University, Canada. His research interests include intelligent transportation systems, machine/deep learning, and autonomous vehicles.

**Zachary Patterson** is Associate Professor in the Department of Geography, Planning and Environment at Concordia University in Montreal. He is also Tier-II Canada Research Chair in Transportation and Land Use Linkages for Regional Sustainability. Dr. Patterson's research has three main thrusts: the use of emerging technologies in data collection, GIS, and statistical analysis. Recent research concentrates on the processing and inference of information from locational data collected from smartphones.

**Bilal Farooq** received B.Eng. degree (2001) from the University of Engineering and Technology, Pakistan, M.Sc. degree (2004) in Computer Science from Lahore University of Management Sciences, Pakistan, and Ph.D. degree (2011) from the University of Toronto, Canada. He is Canada Research Chair in Disruptive Transportation Technologies and Services and an Assistant Professor at Ryerson University. He received Early Researcher Award in Quebec (2014) and Ontario (2018), Canada. His current work focuses on the network and behavioral implications of emerging transportation technologies and services.